\documentclass[twoside, 11pt]{article}

\usepackage[preprint]{jmlr2e}
\usepackage{enumitem}
\usepackage{wrapfig}

\usepackage[utf8]{inputenc}

\usepackage[overlay, absolute]{textpos}

\usepackage{listings}
\usepackage{xcolor}

\definecolor{codegreen}{rgb}{0,0.6,0}
\definecolor{codegray}{rgb}{0.5,0.5,0.5}
\definecolor{codepurple}{rgb}{0.58,0,0.82}
\definecolor{backcolour}{rgb}{0.95,0.95,0.92}

\lstdefinestyle{mystyle}{
    backgroundcolor=\color{backcolour},   
    commentstyle=\color{codegreen},
    keywordstyle=\color{magenta},
    numberstyle=\tiny\color{codegray},
    stringstyle=\color{codepurple},
    basicstyle=\ttfamily\footnotesize,
    breakatwhitespace=false,         
    breaklines=true,                 
    captionpos=b,                    
    keepspaces=true,                 
    numbers=left,                    
    numbersep=2pt,                  
    showspaces=false,                
    showstringspaces=false,
    showtabs=false,                  
    xleftmargin=-0.5cm,
    xrightmargin=0.4cm,
    tabsize=2
}

\usepackage{subcaption}
\usepackage{float} 
\graphicspath{{images/}}

\ShortHeadings{Perception for Autonomous Systems}{Arriaga, Valdenegro-Toro, Muthuraja and Deveramani}
\firstpageno{1}

\begin{document}

\title{Perception for Autonomous Systems (PAZ)}

\author{\name Octavio Arriaga \email arriagac@uni-bremen.de \\
       \addr University of Bremen, Robotics Research Group
       \AND
       \name Matias Valdenegro-Toro \email matias.valdenegro@dfki.de \\
       \addr DFKI - Robotic Innovation Center, Bremen Germany
       \AND
       \name Mohandass Muthuraja \email mohandass@breedingleaders.com \\
       \addr RoBoTec PTC GmbH, Bremen Germany
       \AND
       \name Sushma Devaramani \email sushma.devaramani@inf.h-brs.de \\
       \addr Hochschule Bonn-Rhein Sieg, Sankt Augustin Germany 
       \AND
       \name Frank Kirchner \email frank.kirchner@dfki.de \\
       \addr University of Bremen, Robotics Research Group  \\
       \addr DFKI - Robotic Innovation Center, Bremen Germany
   }

\maketitle

\begin{abstract}
    In this paper we introduce the Perception for Autonomous Systems (PAZ) software library.
    PAZ is a hierarchical perception library that allow users to manipulate multiple levels of abstraction in accordance to their requirements or skill level.
    More specifically, PAZ is divided into three hierarchical levels which we refer to as \verb|pipelines|, \verb|processors|, and \verb|backends|.
     These abstractions allows users to compose functions in a hierarchical modular scheme that can be applied for preprocessing,  data-augmentation, prediction and postprocessing of inputs and outputs of machine learning (ML) models.
     PAZ uses these abstractions to build reusable training and prediction pipelines for multiple robot perception tasks such as: 2D keypoint estimation, 2D object detection, 3D keypoint discovery, 6D pose estimation, emotion classification, face recognition, instance segmentation, and attention mechanisms. 
\end{abstract}


\begin{figure}[h]
    \centering
    \begin{subfigure}[b]{0.16\textwidth}
        \includegraphics[width=\textwidth]{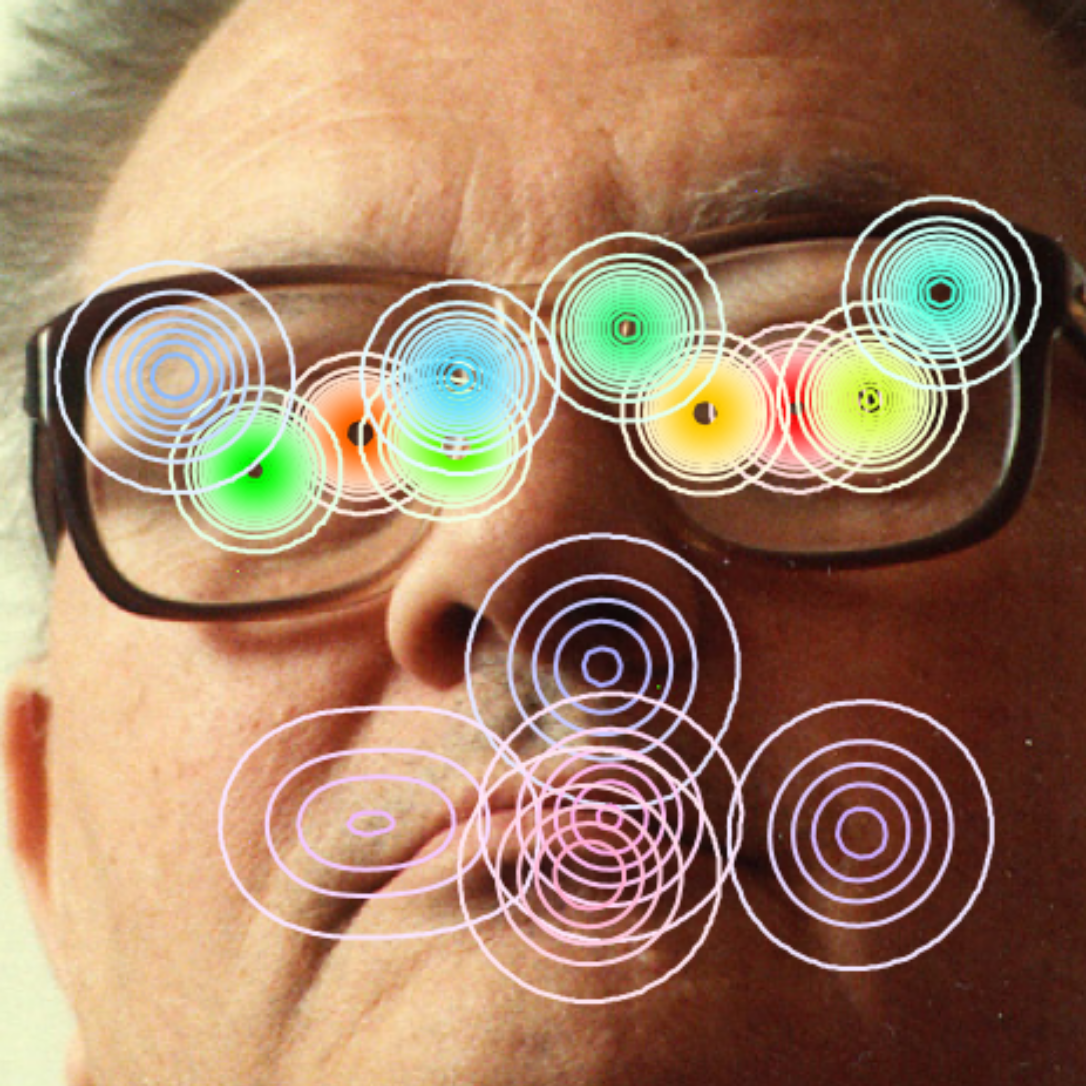}
        \caption{\scriptsize{}}\label{fig:probabilisticKeypoints}
    \end{subfigure}
    \begin{subfigure}[b]{0.16\textwidth}
        \includegraphics[width=\textwidth]{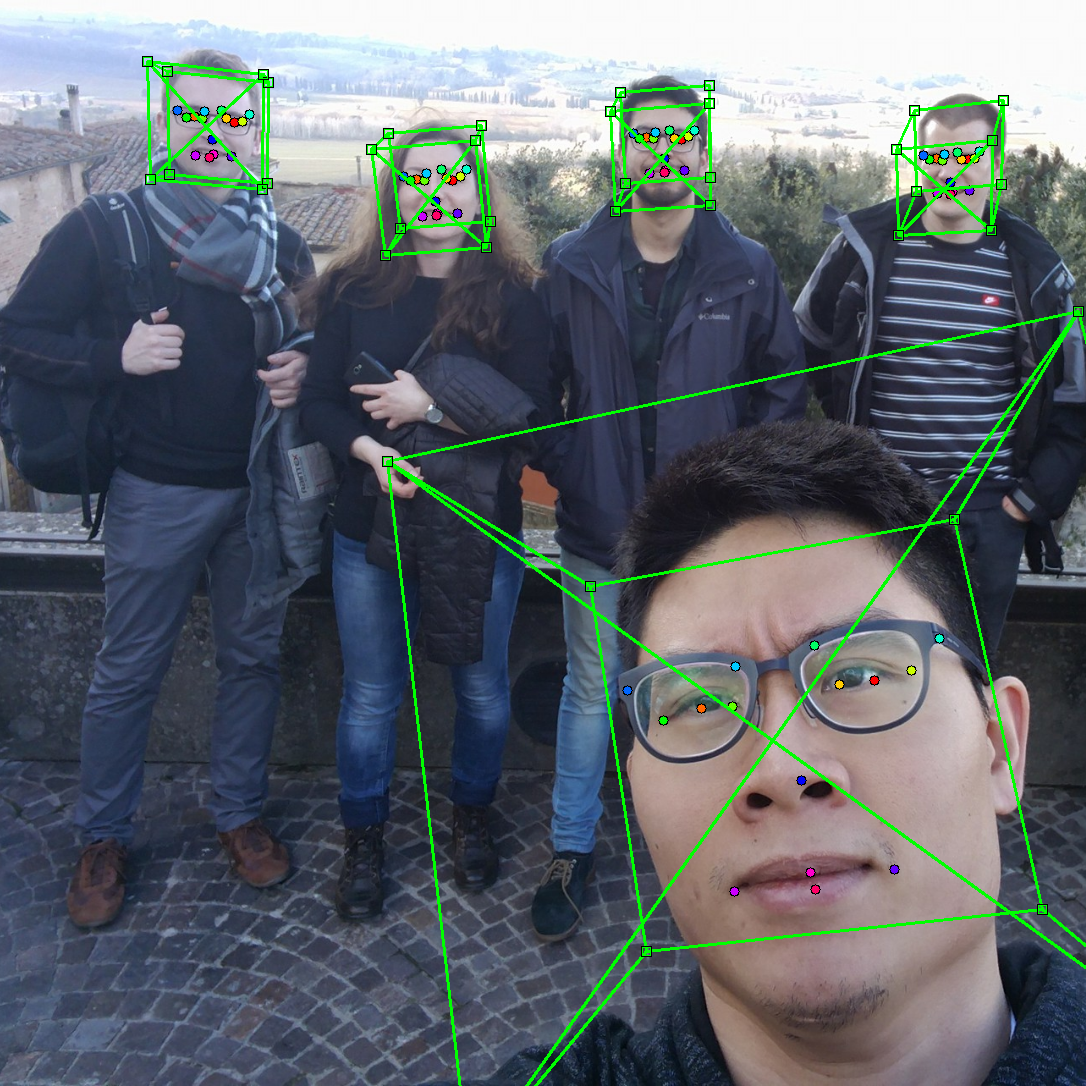}
        \caption{\scriptsize{}}\label{fig:headPoseEstimation}
    \end{subfigure}
    \begin{subfigure}[b]{0.16\textwidth}
        \includegraphics[width=\textwidth]{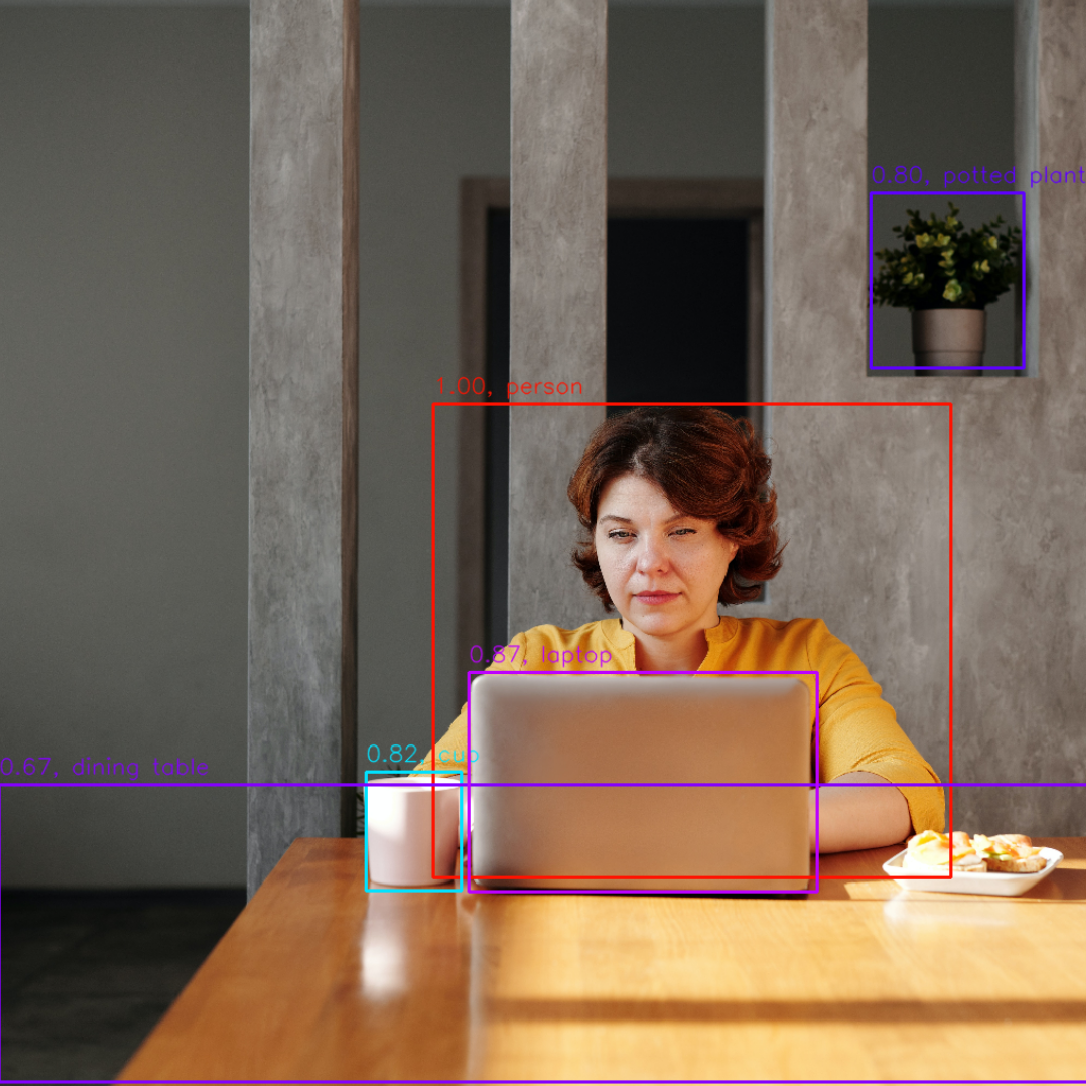}
        \caption{\scriptsize{}}\label{fig:objectDetection}
    \end{subfigure}
    \begin{subfigure}[b]{0.16\textwidth}
        \includegraphics[width=\textwidth]{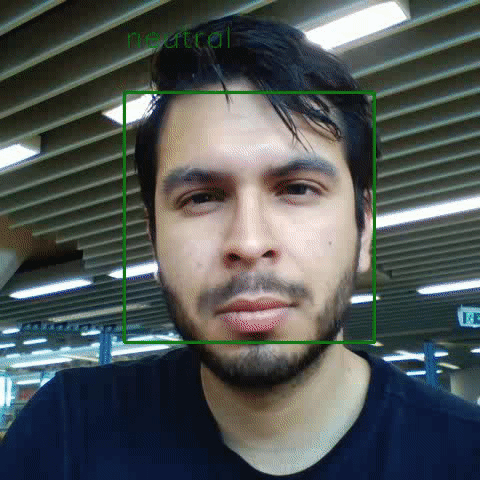}
        \caption{\scriptsize{}}\label{fig:emotionClassification}
    \end{subfigure}
    \begin{subfigure}[b]{0.16\textwidth}
        \includegraphics[width=\textwidth]{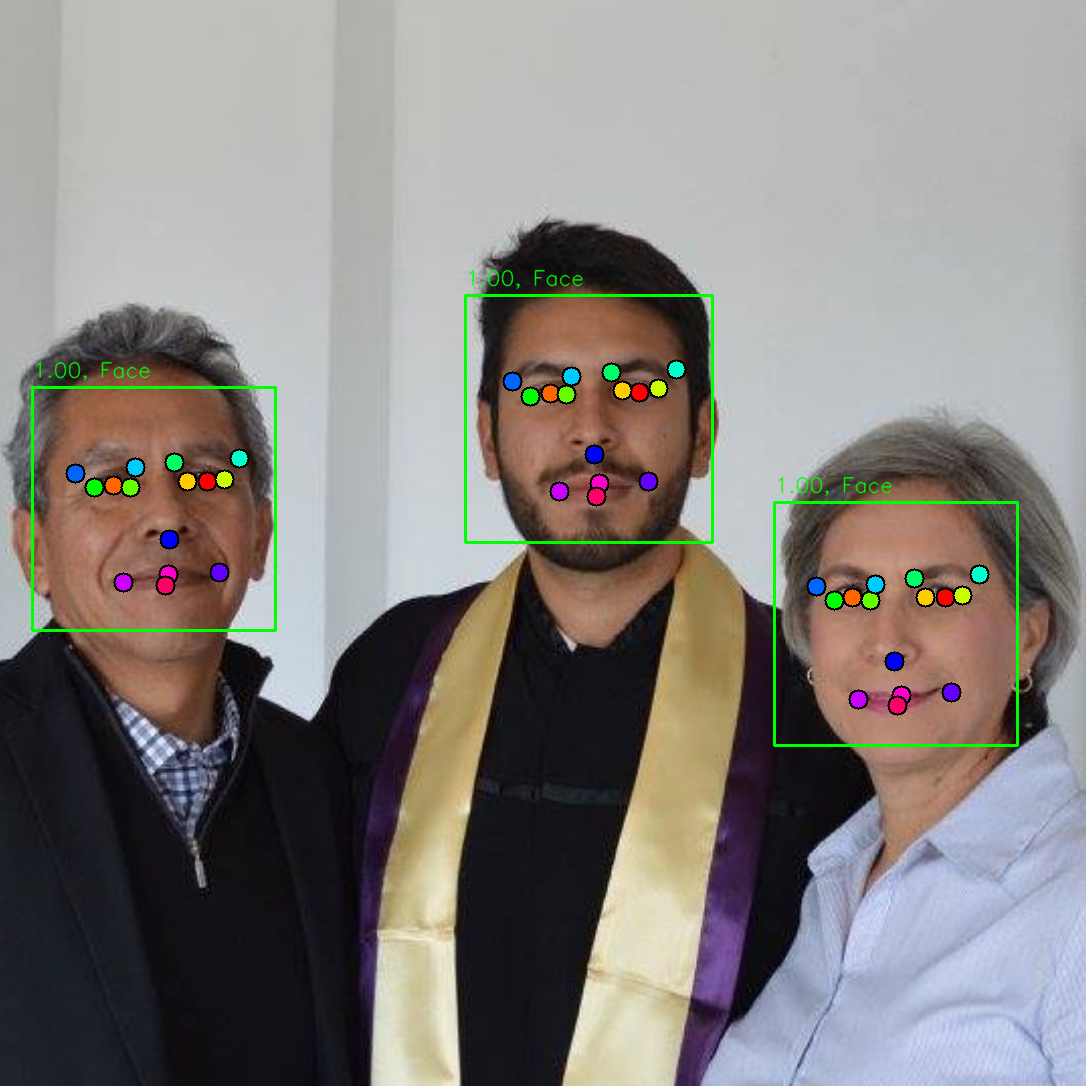}
        \caption{\scriptsize{}}\label{fig:keypointEstimation}
    \end{subfigure}
    \begin{subfigure}[b]{0.16\textwidth}
        \includegraphics[width=\textwidth]{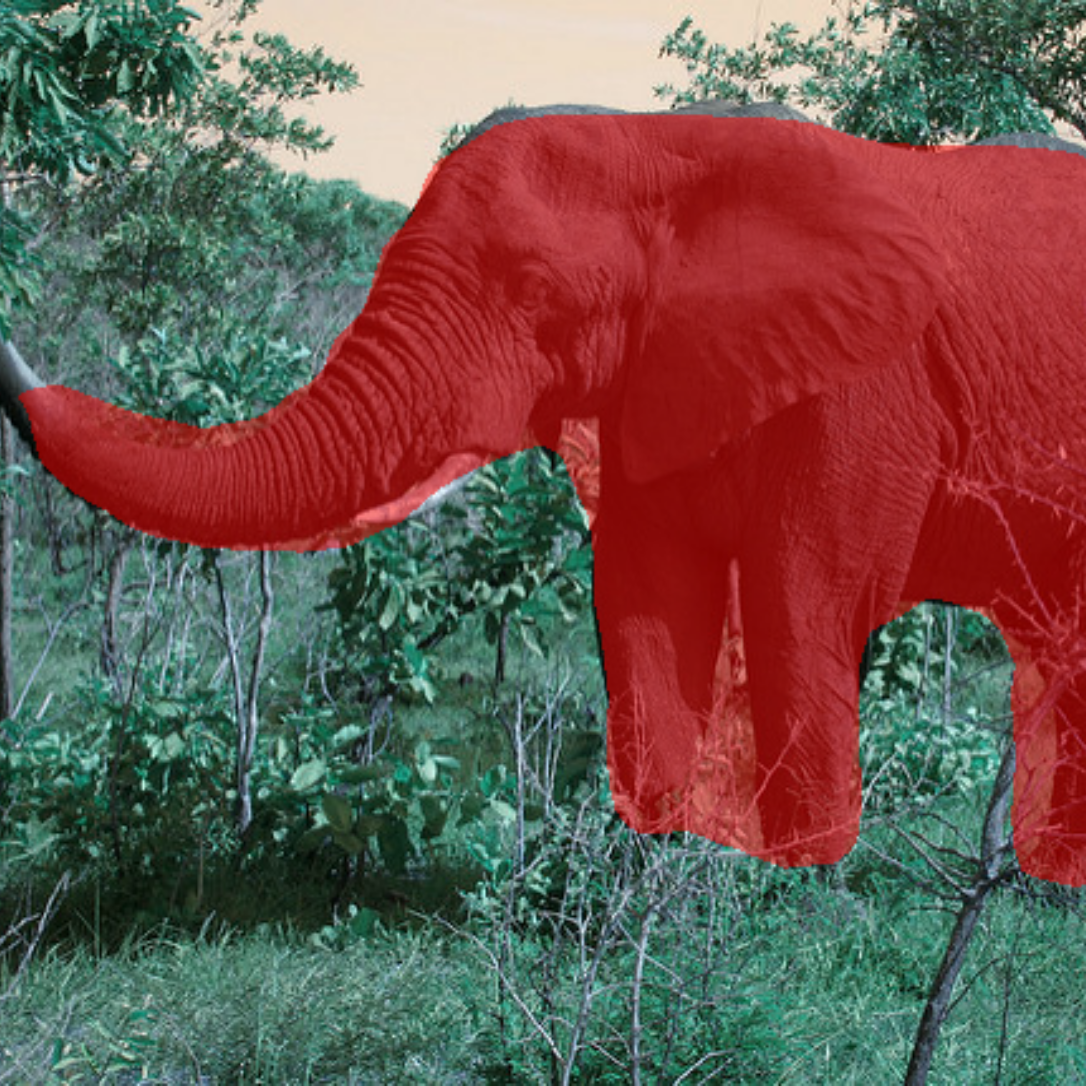}
        \caption{\scriptsize{}}\label{fig:instanceSegmentation}
    \end{subfigure}

    \centering
    \begin{subfigure}[b]{0.16\textwidth}
        \includegraphics[width=\textwidth]{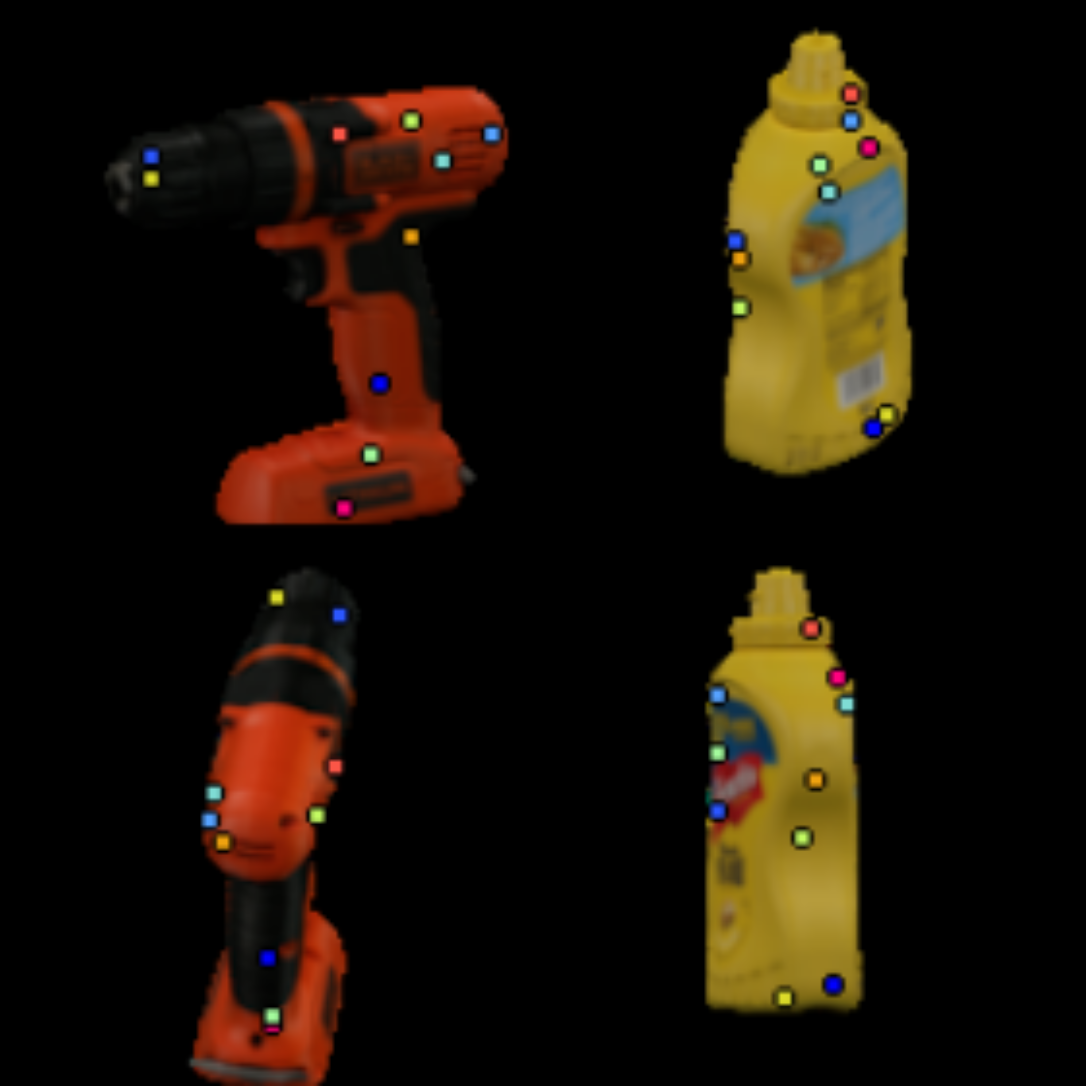}
        \caption{\scriptsize{}}\label{fig:keypointDiscovery}
    \end{subfigure}\label{fig:cifar10_hard}
    \begin{subfigure}[b]{0.16\textwidth}
        \includegraphics[width=\textwidth]{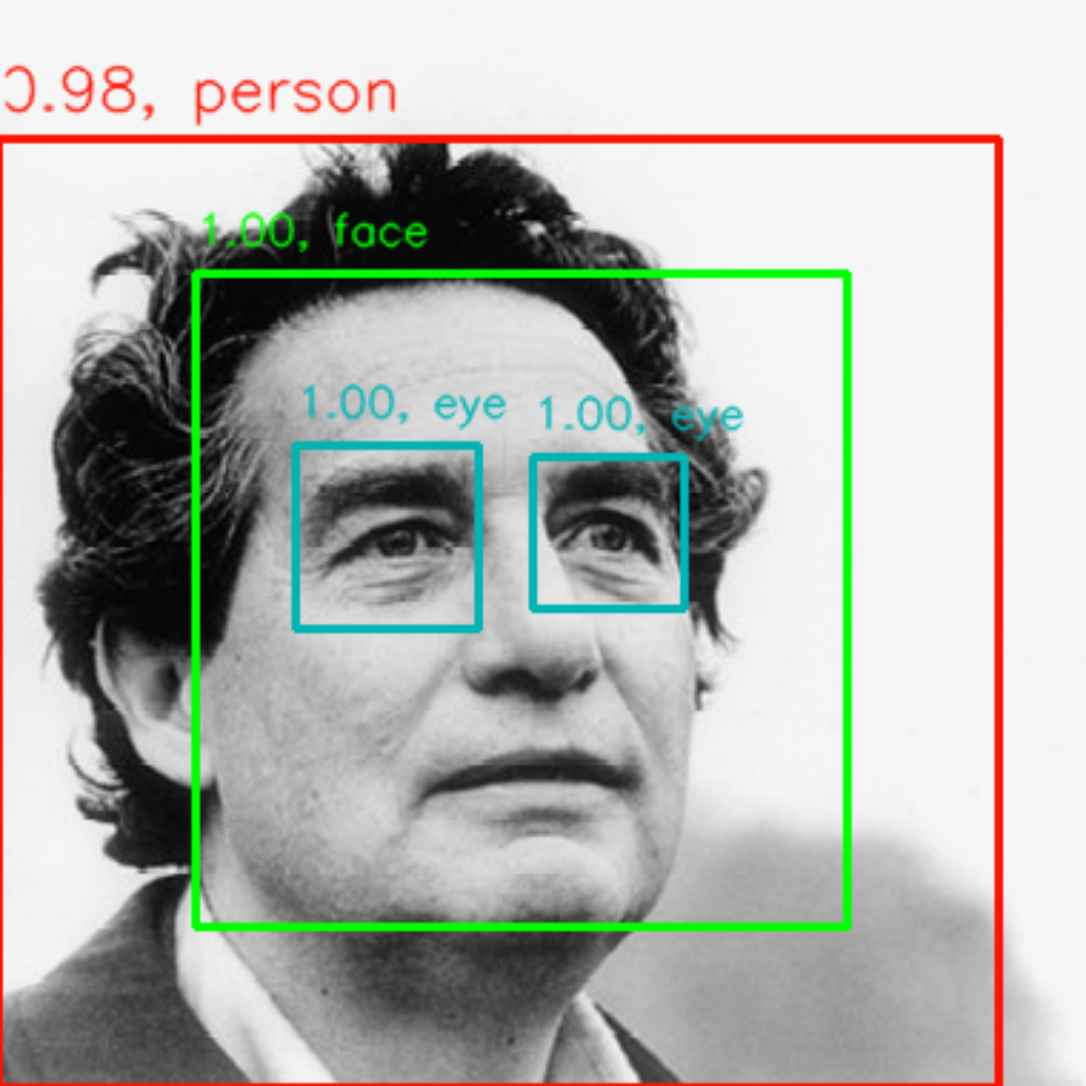}
        \caption{\scriptsize{}}\label{fig:haarCascades}
    \end{subfigure}
    \begin{subfigure}[b]{0.16\textwidth}
        \includegraphics[width=\textwidth]{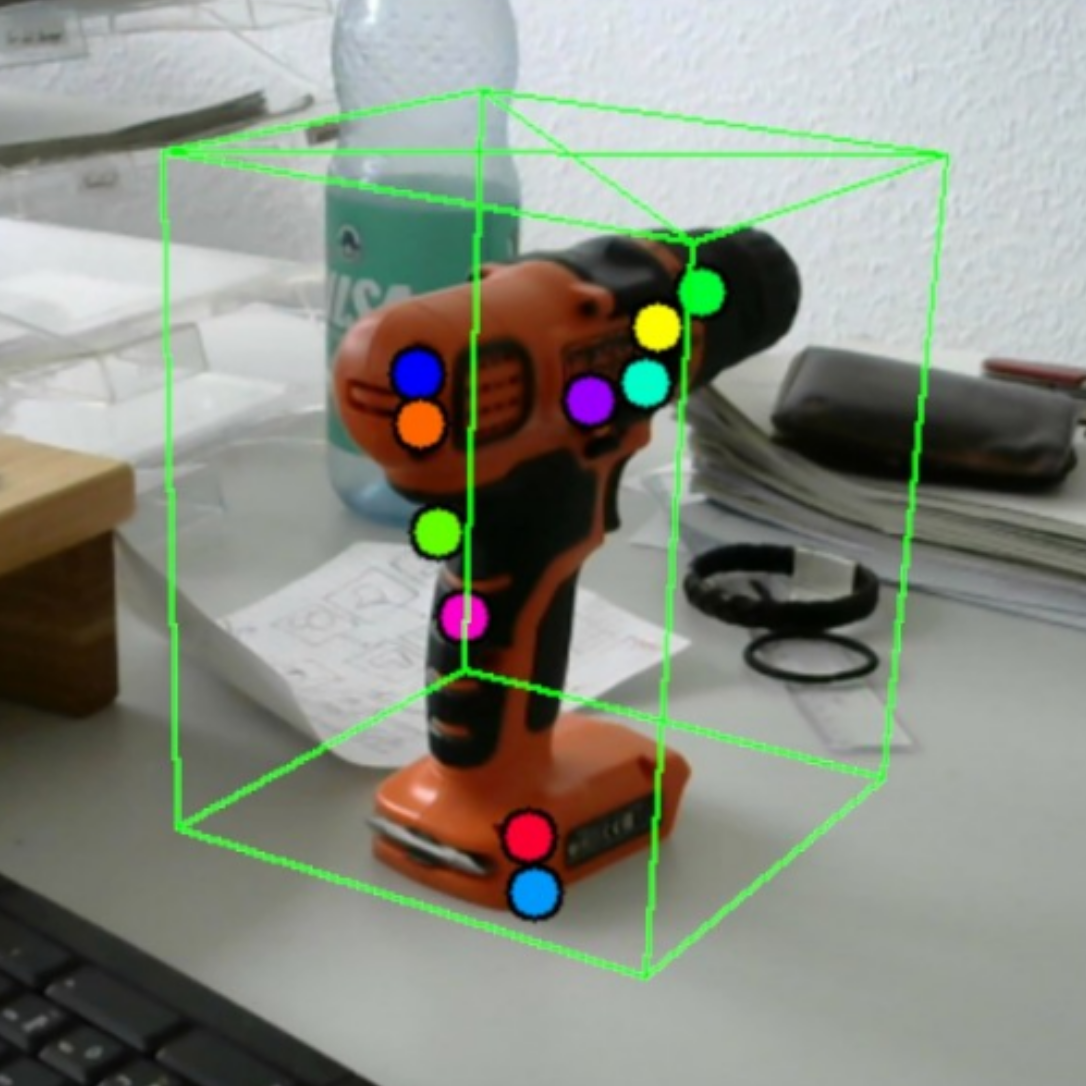}
        \caption{\scriptsize{}}\label{fig:poseEstimation}
    \end{subfigure}
    \begin{subfigure}[b]{0.16\textwidth}
        \includegraphics[width=\textwidth]{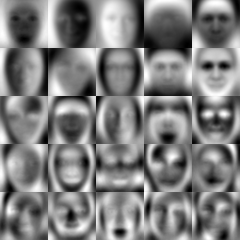}
        \caption{\scriptsize{}}\label{fig:eigenfaces}
    \end{subfigure}
    \begin{subfigure}[b]{0.16\textwidth}
        \includegraphics[width=\textwidth]{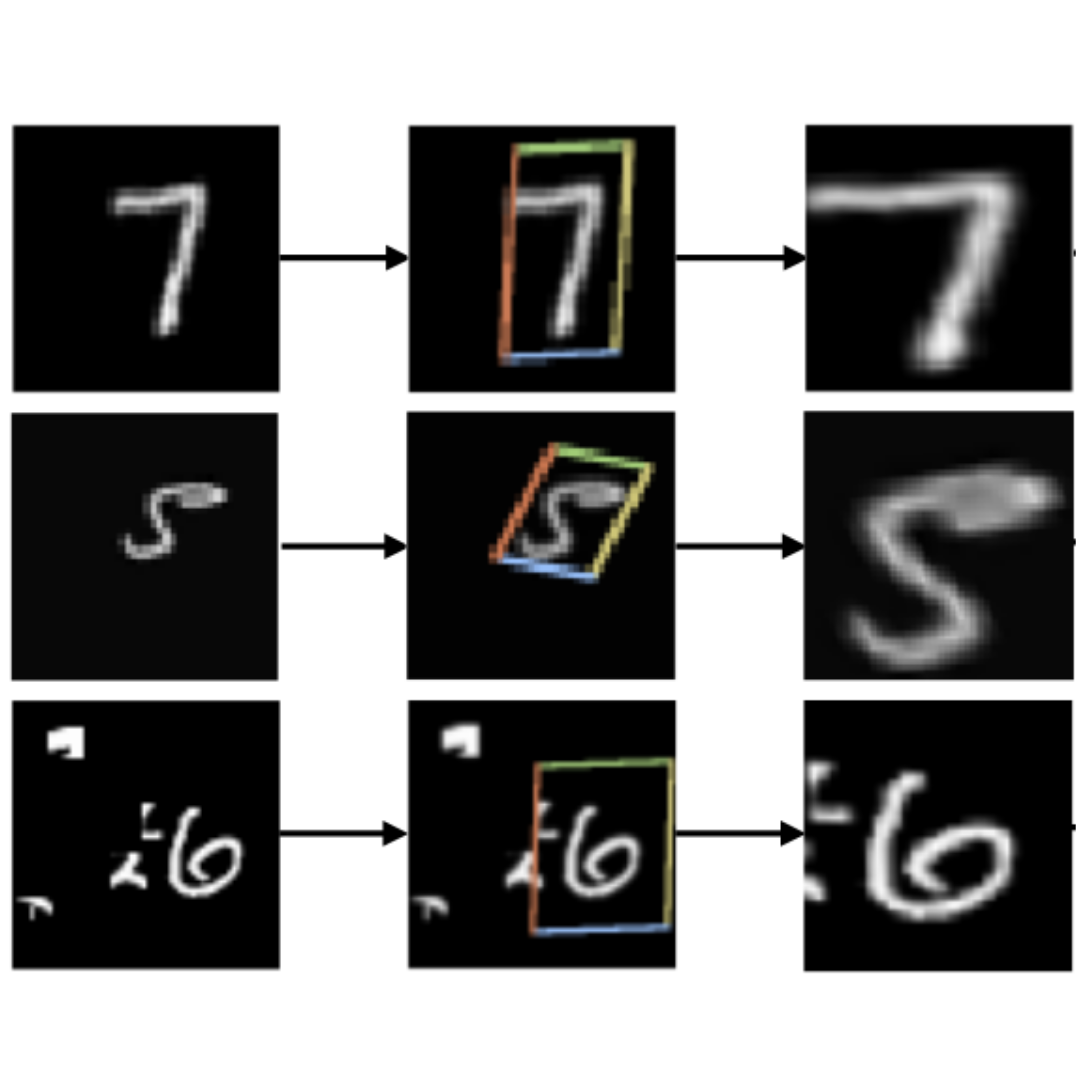}
        \caption{\scriptsize{}}\label{fig:attention}
    \end{subfigure}
    \begin{subfigure}[b]{0.16\textwidth}
        \includegraphics[width=\textwidth]{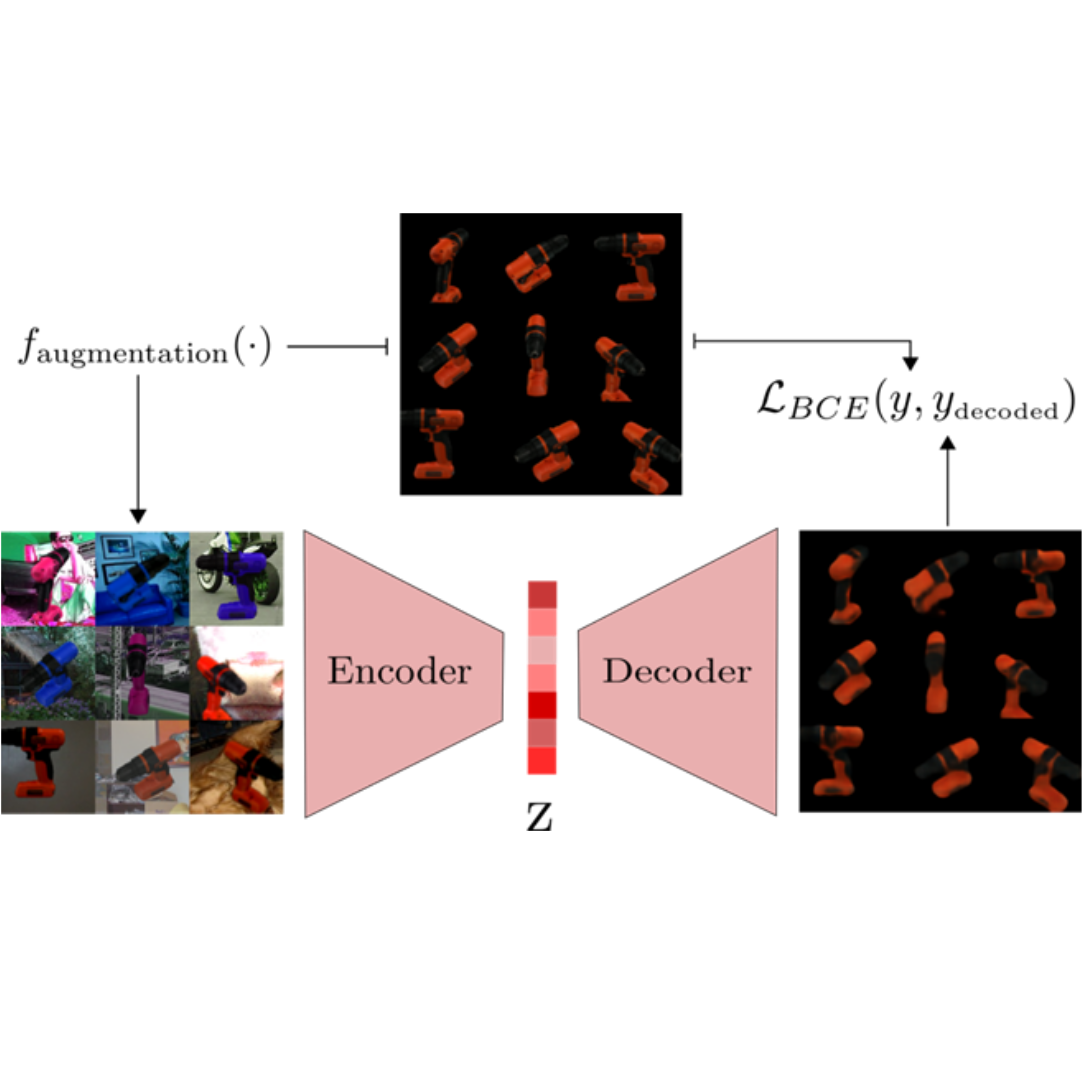}
        \caption{\scriptsize{}}\label{fig:implicitPoseEstimation}
    \end{subfigure}
    \caption{Examples of multiple perception tasks and models implemented using PAZ: \ref{fig:probabilisticKeypoints}) Probabilistic keypoint estimation~\citep{neumann2018tiny} \ref{fig:headPoseEstimation}) 6D head pose estimation \ref{fig:objectDetection}) 2D Object  detection~\citep{liu2016ssd} \ref{fig:emotionClassification}) Emotion clasification~\citep{arriaga2017real} \ref{fig:keypointEstimation}) 2D Keypoint estimation~\citep{wang2020deep} \ref{fig:instanceSegmentation}) Instance segmentation~\citep{he2017mask} \ref{fig:keypointDiscovery}) Keypoint discovery~\citep{suwajanakorn2018discovery} \ref{fig:haarCascades}) Haar Cascades~\citep{viola2001robust} \ref{fig:poseEstimation}) 6D Pose estimation \ref{fig:eigenfaces}) Face recognition~\citep{turk1991eigenfaces} \ref{fig:attention}) Attention~\citep{jaderberg2015spatial} \ref{fig:implicitPoseEstimation}) Implicit pose estimation~\citep{sundermeyer2018implicit}.}\label{fig:PAZExamples}
\end{figure}\label{fig:PAZExamples}

\section{Introduction}
    Current research trends in machine learning and computer vision indicate that state-of-the-art models in the coming years would most likely include the use of differentiable libraries and deep learning~\citep{Goodfellow-et-al-2016}.
    Furthermore, deep learning has now been consolidated as an incremental discipline in which we hypothesize that small additive changes must also represent small incremental modifications to their software description.
    Additionally, the term \textit{design stamina hypothesis} is used in software engineering to describe the capacity of software to quickly develop additional functionalities given that it contains an appropriate set of internal tools and abstractions~\citep{fowler2018refactoring}.
    Thus, one of the main goals of PAZ is to create internal software structures that satisfy the \textit{design stamina hypothesis} for perceptual algorithms.
    We corroborate the successful direction of PAZ by using our hierarchical abstractions to build reusable components for multiple robot perception algorithms in multiple domains.
    More specifically, PAZ was used for creating complete training and inference pipelines for all tasks and models displayed in Figure~\ref{fig:PAZExamples}.

    \section{Related software}
    Similar software libraries are Detectron2~\citep{wu2019detectron2} the Object detection API from Tensorflow~\citep{huang2017speed} and MMDetection~\citep{chen2019mmdetection}.
    Detectron2 focuses in extending a single model, Mask-RCNN, while the Object Detection API and MMDetection focus primarily in optimizing models on the single task of 2D object detection.
    On the other hand, as shown in Figure~\ref{fig:PAZExamples}, PAZ focuses in extending multiple models across a diverse set of perception tasks.
    This broad generality of tasks and models is possible due to our hierarchical-API, which allows users to re-use and construct entirely new functions in a modular scheme.

    \section{Hierarchical abstractions}    
    In this section we review the main components of each of our hierarchical levels and their corresponding software abstractions.
    One important consideration to be made is that while we encourage the user to use our abstractions we don't necessarily impose them.
    Thus, a PAZ user can use any of it's functionality at any level without necessarily subscribing to a specific API.
    \subsection{High-level:}
    Our highest API level, \verb|pipelines|, contains application-ready functions for 2D object detection, 2D keypoint estimation, 6D pose estimation, emotion classification, data-augmentation and image pre-processing.
    Our API allows the user to quickly instantiate out-of-the-box functions that can be applied directly to an image.
    Listing 1 contains an example on how to instantiate a single-shot object detector from PAZ and apply directly to an image for an end-to-end prediction. 

\begin{minipage}[t]{0.47\textwidth}
\begin{lstlisting}[language=Python, caption=High-level function]
from paz.pipelines import SSD512COCO

detect = SSD512COCO()

# apply directly to a numpy-array
inferences = detect(image)

# ``detect'' performs preprocessing
# and postprocessing steps to form
# an end-to-end prediction function
\end{lstlisting}\label{lst:highLevel}
\end{minipage}\hspace{0.5cm}
\begin{minipage}[t]{0.47\textwidth}
\begin{lstlisting}[language=Python, caption=Mid-level construction]
from paz import processors as pr

augment = pr.SequentialProcessor()
augment.add(pr.RandomContrast())
augment.add(pr.RandomBrightness())
augment.add(pr.RandomSaturation())
augment.add(pr.RandomHue())

# apply it as a python function
image = augment(image)
\end{lstlisting}
\end{minipage}



    
    \subsection{Mid-level}
    The high-level API is useful for rapidly creating applications; however, it might not be flexible enough for the user's specific purposes.
    Therefore, PAZ builds high-level functions using our a mid-level API which allows the user to modify or extend existing pipelines.
    The abstraction for this mid-level is referred to as a \verb|Processor|\footnote{The name \textit{processor} describes the capacity to create both pre-processing and post-processing functions.}.
    Processors are meant to perform small computations that can be re-used in other applications or entirely new algorithms.
    PAZ includes the \verb|SequentialProcessor| abstraction to sequentially apply processors to a set of inputs\footnote{Similar abstractions have proven useful in the context of deep learning.
    Specifically in the sequential application of \textit{layers} in Keras~\citep{chollet2015keras}.}.
    In Listing 2 we create a simple data-augmentation pipeline for image classification.
    Our sequential API reveals some of the flexibility and reusability of PAZ.
    If for example a user wishes to input a dictionary, or to add a new data augmentation function or a normalization operation one would only need to add a new processor.
    Furthemore, PAZ provides an abstract template class for creating any custom new logic.
    However, an important consideration that we would like to make is that the user can pass any python function to a \verb|pr.Sequential| pipeline, and is not constrained to use our \verb|Processor| base class.
    Another relevant aspect of our API is that it clearly depicts the processing steps of our data into well separated modules; thus, PAZ creates a programming bias to distribute computation into multiple simple functions.
    This allows users with limited experience either with programming or with a specific new algorithm to easily adapt, debug, or understand any aspect of it's computation.
    We specifically consider this of great importance given the state of many ML research projects, that often prove difficult to continue their improvement.

    \subsection{Low-level}
    Processors allow us to easily compose, compress and extract away parameters of functions; however, most processors are build using our low-level API (\verb|backend|).
    Our \textit{backend} modules are found in: \verb|backend.boxes|, \verb|backend.camera|, \verb|backend.image|, \verb|backend.keypoints| and \verb|backend.quaternion|.
    Each of these modules is meant to be expanded or entirely replaced without affecting the functionality of the higher levels.
    For example, if a camera contains it's own software API, one could wrap this camera-specific API with our \verb|backend.camera.Camera| fields and methods in order to re-use our own specific camera utilities such as real-time prediction visualization or real-time prediction video-recording.

    Furthermore, one could re-use any low-level functions without ever subscribing to our mid-level or high-level APIs or functionalities.
    In this case one uses PAZ as a collection of small modules without any meta-function construction.
    We believe that a user should have the possibility to choose which paradigms fits better to their case, ability or style; thus, we provide functionality and resources for each of these levels.

    \section{Additional functionality}

    \begin{itemize}[leftmargin=*]
        \item[] \textbf{Built-in messages:}
            PAZ includes built-in messages of common prediction types made in perceptual systems.
            These built-in messages include \verb|Box2D|, \verb|Pose6D| and \verb|Keypoints3D|.
            These types allow PAZ users to have an easier data exchange with other robotic frameworks such as ROS~\citep{quigley2009ros} or ROCK~\citep{joyeux2013rock} without having to install any additional software.

        \item[] \textbf{Datasets:}
            PAZ provides a common interface to load multiple datasets related to object detection, image segmentation and image classification.
            The available datasets within PAZ are OpenImages~\citep{kuznetsova2018open}, COCO~\citep{lin2014microsoft}, VOC~\citep{everingham2010pascal}, YCB-Video~\citep{xiang2017posecnn}, FAT~\citep{tremblay2018deep}, FERPlus~\citep{BarsoumICMI2016} and FER2013~\citep{goodfellow2013challenges}.

        \item[] \textbf{Automatic batch dispatching:}
            Once a dataset has been loaded we can pass it to our batch dispatcher class (\verb|SequenceProcessing|), along with any built-in or custom function for prepreprocessing or data augmentation.
            The batch dispatcher class instantiates a generator that is ready to be used directly with a \verb|model.fit| scheme.

    \end{itemize}

    \section{Software engineering}
    PAZ has only three dependencies: Tensorflow \citep{Tensorflow2015}, OpenCV \citep{Bradski2000opencv} and Numpy \citep{Van2011numpy}.
    Furthermore, it has continuous integration (CI) in multiple python versions (python 3.5, 3.6, 3.7 and 3.8).
    PAZ has unit-tests for all high-level application functions along with most of the major \verb|backend| modules, and it currently has a test-coverage of 47\%.
    Additionally, PAZ has automatic documentation generation directly from documentation strings.

    \section{Conclusions and future work}
    We presented a modular hierarchical software library for perceptual systems.
    We validated its applicability by creating training and prediction pipelines on a wide range of tasks and algorithms.
    Further directions and improvements of this software library include increasing the perceptual tasks, increasing the test coverage, and extending our backend functionality.

    \section{Acknowledgments}
    We would like to thank Alexander Fabisch for his insightful evaluation and thorough revision of our software.
    Furthermore, we would like to thank the internal DFKI software board for their evaluation metrics and helpful discussions.
    This work was supported through two grants of the German Federal Ministry of Economics and Energy during the Projects TransFIT and KiMMI-SF [BMWi, FKZ 50 RA 1703, and FKZ 50 RA 2022].

\bibliography{references}

\begin{thebibliography}{28}
\providecommand{\natexlab}[1]{#1}
\providecommand{\url}[1]{\texttt{#1}}
\expandafter\ifx\csname urlstyle\endcsname\relax
  \providecommand{\doi}[1]{doi: #1}\else
  \providecommand{\doi}{doi: \begingroup \urlstyle{rm}\Url}\fi

\bibitem[Abadi et~al.(2015)Abadi, Agarwal, Barham, Brevdo, Chen, Citro,
  Corrado, Davis, Dean, Devin, Ghemawat, Goodfellow, Harp, Irving, Isard, Jia,
  Jozefowicz, Kaiser, Kudlur, Levenberg, Man\'{e}, Monga, Moore, Murray, Olah,
  Schuster, Shlens, Steiner, Sutskever, Talwar, Tucker, Vanhoucke, Vasudevan,
  Vi\'{e}gas, Vinyals, Warden, Wattenberg, Wicke, Yu, and
  Zheng]{Tensorflow2015}
Mart\'{\i}n Abadi, Ashish Agarwal, Paul Barham, Eugene Brevdo, Zhifeng Chen,
  Craig Citro, Greg~S. Corrado, Andy Davis, Jeffrey Dean, Matthieu Devin,
  Sanjay Ghemawat, Ian Goodfellow, Andrew Harp, Geoffrey Irving, Michael Isard,
  Yangqing Jia, Rafal Jozefowicz, Lukasz Kaiser, Manjunath Kudlur, Josh
  Levenberg, Dandelion Man\'{e}, Rajat Monga, Sherry Moore, Derek Murray, Chris
  Olah, Mike Schuster, Jonathon Shlens, Benoit Steiner, Ilya Sutskever, Kunal
  Talwar, Paul Tucker, Vincent Vanhoucke, Vijay Vasudevan, Fernanda Vi\'{e}gas,
  Oriol Vinyals, Pete Warden, Martin Wattenberg, Martin Wicke, Yuan Yu, and
  Xiaoqiang Zheng.
\newblock {TensorFlow}: Large-scale machine learning on heterogeneous systems,
  2015.
\newblock URL \url{https://www.tensorflow.org/}.
\newblock Software available from tensorflow.org.

\bibitem[Arriaga et~al.(2017)Arriaga, Valdenegro-Toro, and
  Pl{\"o}ger]{arriaga2017real}
Octavio Arriaga, Matias Valdenegro-Toro, and Paul Pl{\"o}ger.
\newblock Real-time convolutional neural networks for emotion and gender
  classification.
\newblock \emph{arXiv preprint arXiv:1710.07557}, 2017.

\bibitem[Barsoum et~al.(2016)Barsoum, Zhang, Canton~Ferrer, and
  Zhang]{BarsoumICMI2016}
Emad Barsoum, Cha Zhang, Cristian Canton~Ferrer, and Zhengyou Zhang.
\newblock Training deep networks for facial expression recognition with
  crowd-sourced label distribution.
\newblock In \emph{ACM International Conference on Multimodal Interaction
  (ICMI)}, 2016.

\bibitem[Bradski(2000)]{Bradski2000opencv}
Gary Bradski.
\newblock The opencv library.
\newblock \emph{Dr Dobb's J. Software Tools}, 25:\penalty0 120--125, 2000.

\bibitem[Chen et~al.(2019)Chen, Wang, Pang, Cao, Xiong, Li, Sun, Feng, Liu, Xu,
  et~al.]{chen2019mmdetection}
Kai Chen, Jiaqi Wang, Jiangmiao Pang, Yuhang Cao, Yu~Xiong, Xiaoxiao Li,
  Shuyang Sun, Wansen Feng, Ziwei Liu, Jiarui Xu, et~al.
\newblock Mmdetection: Open mmlab detection toolbox and benchmark.
\newblock \emph{arXiv preprint arXiv:1906.07155}, 2019.

\bibitem[Chollet et~al.(2015)]{chollet2015keras}
Fran\c{c}ois Chollet et~al.
\newblock Keras.
\newblock \url{https://keras.io}, 2015.

\bibitem[Everingham et~al.(2010)Everingham, Van~Gool, Williams, Winn, and
  Zisserman]{everingham2010pascal}
Mark Everingham, Luc Van~Gool, Christopher~KI Williams, John Winn, and Andrew
  Zisserman.
\newblock The pascal visual object classes (voc) challenge.
\newblock \emph{International journal of computer vision}, 88\penalty0
  (2):\penalty0 303--338, 2010.

\bibitem[Fowler(2018)]{fowler2018refactoring}
Martin Fowler.
\newblock \emph{Refactoring: improving the design of existing code}.
\newblock Addison-Wesley Professional, 2018.

\bibitem[Goodfellow et~al.(2016)Goodfellow, Bengio, and
  Courville]{Goodfellow-et-al-2016}
Ian Goodfellow, Yoshua Bengio, and Aaron Courville.
\newblock \emph{Deep Learning}.
\newblock MIT Press, 2016.
\newblock \url{http://www.deeplearningbook.org}.

\bibitem[Goodfellow et~al.(2013)Goodfellow, Erhan, Carrier, Courville, Mirza,
  Hamner, Cukierski, Tang, Thaler, Lee, et~al.]{goodfellow2013challenges}
Ian~J Goodfellow, Dumitru Erhan, Pierre~Luc Carrier, Aaron Courville, Mehdi
  Mirza, Ben Hamner, Will Cukierski, Yichuan Tang, David Thaler, Dong-Hyun Lee,
  et~al.
\newblock Challenges in representation learning: A report on three machine
  learning contests.
\newblock In \emph{International conference on neural information processing},
  pages 117--124. Springer, 2013.

\bibitem[He et~al.(2017)He, Gkioxari, Doll{\'a}r, and Girshick]{he2017mask}
Kaiming He, Georgia Gkioxari, Piotr Doll{\'a}r, and Ross Girshick.
\newblock Mask r-cnn.
\newblock In \emph{Proceedings of the IEEE international conference on computer
  vision}, pages 2961--2969, 2017.

\bibitem[Huang et~al.(2017)Huang, Rathod, Sun, Zhu, Korattikara, Fathi,
  Fischer, Wojna, Song, Guadarrama, et~al.]{huang2017speed}
Jonathan Huang, Vivek Rathod, Chen Sun, Menglong Zhu, Anoop Korattikara,
  Alireza Fathi, Ian Fischer, Zbigniew Wojna, Yang Song, Sergio Guadarrama,
  et~al.
\newblock Speed/accuracy trade-offs for modern convolutional object detectors.
\newblock In \emph{Proceedings of the IEEE conference on computer vision and
  pattern recognition}, pages 7310--7311, 2017.

\bibitem[Jaderberg et~al.(2015)Jaderberg, Simonyan, Zisserman,
  et~al.]{jaderberg2015spatial}
Max Jaderberg, Karen Simonyan, Andrew Zisserman, et~al.
\newblock Spatial transformer networks.
\newblock In \emph{Advances in neural information processing systems}, pages
  2017--2025, 2015.

\bibitem[Joyeux(2013)]{joyeux2013rock}
Sylvain Joyeux.
\newblock Rock: the robot construction kit, 2013.

\bibitem[Kuznetsova et~al.(2018)Kuznetsova, Rom, Alldrin, Uijlings, Krasin,
  Pont-Tuset, Kamali, Popov, Malloci, Duerig, et~al.]{kuznetsova2018open}
Alina Kuznetsova, Hassan Rom, Neil Alldrin, Jasper Uijlings, Ivan Krasin, Jordi
  Pont-Tuset, Shahab Kamali, Stefan Popov, Matteo Malloci, Tom Duerig, et~al.
\newblock The open images dataset v4: Unified image classification, object
  detection, and visual relationship detection at scale.
\newblock \emph{arXiv preprint arXiv:1811.00982}, 2018.

\bibitem[Lin et~al.(2014)Lin, Maire, Belongie, Hays, Perona, Ramanan,
  Doll{\'a}r, and Zitnick]{lin2014microsoft}
Tsung-Yi Lin, Michael Maire, Serge Belongie, James Hays, Pietro Perona, Deva
  Ramanan, Piotr Doll{\'a}r, and C~Lawrence Zitnick.
\newblock Microsoft coco: Common objects in context.
\newblock In \emph{European conference on computer vision}, pages 740--755.
  Springer, 2014.

\bibitem[Liu et~al.(2016)Liu, Anguelov, Erhan, Szegedy, Reed, Fu, and
  Berg]{liu2016ssd}
Wei Liu, Dragomir Anguelov, Dumitru Erhan, Christian Szegedy, Scott Reed,
  Cheng-Yang Fu, and Alexander~C Berg.
\newblock Ssd: Single shot multibox detector.
\newblock In \emph{European conference on computer vision}, pages 21--37.
  Springer, 2016.

\bibitem[Neumann and Vedaldi(2018)]{neumann2018tiny}
Luk{\'a}{\v{s}} Neumann and Andrea Vedaldi.
\newblock Tiny people pose.
\newblock In \emph{Asian Conference on Computer Vision}, pages 558--574.
  Springer, 2018.

\bibitem[Quigley et~al.(2009)Quigley, Conley, Gerkey, Faust, Foote, Leibs,
  Wheeler, and Ng]{quigley2009ros}
Morgan Quigley, Ken Conley, Brian Gerkey, Josh Faust, Tully Foote, Jeremy
  Leibs, Rob Wheeler, and Andrew~Y Ng.
\newblock Ros: an open-source robot operating system.
\newblock In \emph{ICRA workshop on open source software}, volume~3, page~5.
  Kobe, Japan, 2009.

\bibitem[Sundermeyer et~al.(2018)Sundermeyer, Marton, Durner, Brucker, and
  Triebel]{sundermeyer2018implicit}
Martin Sundermeyer, Zoltan-Csaba Marton, Maximilian Durner, Manuel Brucker, and
  Rudolph Triebel.
\newblock Implicit 3d orientation learning for 6d object detection from rgb
  images.
\newblock In \emph{Proceedings of the European Conference on Computer Vision
  (ECCV)}, pages 699--715, 2018.

\bibitem[Suwajanakorn et~al.(2018)Suwajanakorn, Snavely, Tompson, and
  Norouzi]{suwajanakorn2018discovery}
Supasorn Suwajanakorn, Noah Snavely, Jonathan~J Tompson, and Mohammad Norouzi.
\newblock Discovery of latent 3d keypoints via end-to-end geometric reasoning.
\newblock In \emph{Advances in neural information processing systems}, pages
  2059--2070, 2018.

\bibitem[Tremblay et~al.(2018)Tremblay, To, Sundaralingam, Xiang, Fox, and
  Birchfield]{tremblay2018deep}
Jonathan Tremblay, Thang To, Balakumar Sundaralingam, Yu~Xiang, Dieter Fox, and
  Stan Birchfield.
\newblock Deep object pose estimation for semantic robotic grasping of
  household objects.
\newblock \emph{arXiv preprint arXiv:1809.10790}, 2018.

\bibitem[Turk and Pentland(1991)]{turk1991eigenfaces}
Matthew Turk and Alex Pentland.
\newblock Eigenfaces for recognition.
\newblock \emph{Journal of cognitive neuroscience}, 3\penalty0 (1):\penalty0
  71--86, 1991.

\bibitem[Van Der~Walt et~al.(2011)Van Der~Walt, Colbert, and
  Varoquaux]{Van2011numpy}
Stefan Van Der~Walt, S~Chris Colbert, and Gael Varoquaux.
\newblock The numpy array: a structure for efficient numerical computation.
\newblock \emph{Computing in Science \& Engineering}, 13\penalty0 (2):\penalty0
  22, 2011.

\bibitem[Viola and Jones(2001)]{viola2001robust}
Paul Viola and Michael Jones.
\newblock Robust real-time face detection.
\newblock In \emph{null}, page 747. IEEE, 2001.

\bibitem[Wang et~al.(2020)Wang, Sun, Cheng, Jiang, Deng, Zhao, Liu, Mu, Tan,
  Wang, et~al.]{wang2020deep}
Jingdong Wang, Ke~Sun, Tianheng Cheng, Borui Jiang, Chaorui Deng, Yang Zhao,
  Dong Liu, Yadong Mu, Mingkui Tan, Xinggang Wang, et~al.
\newblock Deep high-resolution representation learning for visual recognition.
\newblock \emph{IEEE transactions on pattern analysis and machine
  intelligence}, 2020.

\bibitem[Wu et~al.(2019)Wu, Kirillov, Massa, Lo, and
  Girshick]{wu2019detectron2}
Yuxin Wu, Alexander Kirillov, Francisco Massa, Wan-Yen Lo, and Ross Girshick.
\newblock Detectron2, 2019.

\bibitem[Xiang et~al.(2017)Xiang, Schmidt, Narayanan, and
  Fox]{xiang2017posecnn}
Yu~Xiang, Tanner Schmidt, Venkatraman Narayanan, and Dieter Fox.
\newblock Posecnn: A convolutional neural network for 6d object pose estimation
  in cluttered scenes.
\newblock \emph{arXiv preprint arXiv:1711.00199}, 2017.

\end{thebibliography}

\end{document}